\documentclass{article} 

\usepackage{amsmath,amsfonts,bm}

\def\eqref#1{equation~\ref{#1}}

\def\1{\bm{1}}

\DeclareMathAlphabet{\mathsfit}{\encodingdefault}{\sfdefault}{m}{sl}
\SetMathAlphabet{\mathsfit}{bold}{\encodingdefault}{\sfdefault}{bx}{n}

\newcommand{\E}{\mathbb{E}}

\newcommand{\R}{\mathbb{R}}

\DeclareMathOperator*{\argmin}{arg\,min}

\DeclareMathOperator{\sign}{sign}

\usepackage{hyperref,natbib}
\usepackage{url}
\usepackage[utf8]{inputenc} 
\usepackage[T1]{fontenc}    
\usepackage{hyperref}       
\usepackage{url}            
\usepackage{booktabs}       
\usepackage{amsfonts}       
\usepackage{nicefrac}       
\usepackage{microtype}      
\usepackage{xcolor}         

\usepackage{xcolor}
\usepackage{algpseudocode,algorithmicx,algorithm,graphicx,geometry}
\usepackage{amsmath,amsthm,amssymb, bm}

\newtheorem{ppty}{Property}
\newtheorem{definition}{Definition}
\usepackage[caption=false]{subfig}
\usepackage{wrapfig}
\usepackage[normalem]{ulem}

\newcommand{\ccp}{\mathcal{P}}

\newcommand{\intr}{\textnormal{int}}
\newcommand{\diag}{\textnormal{diag}}
\newcommand\norm[1]{\lVert#1\rVert}

\title{RbX: Region-based explanations of prediction models}

\author{Ismael Lemhadri, Harrison H. Li, and Trevor Hastie \\
}

\begin{document}

\maketitle

\begin{abstract}
We introduce region-based explanations (RbX), 
a novel, model-agnostic method 
to generate local explanations of scalar outputs from a black-box prediction model 
using only query access. 
RbX is based on a greedy algorithm for building a convex polytope 
that approximates a region of feature space 
where model predictions are close to the prediction at some target point. 
This region is fully specified by the user on the scale of the predictions, 
rather than on the scale of the features. 
The geometry of this polytope --- 
specifically the change in each coordinate necessary to escape the polytope --- 
quantifies the local sensitivity of the predictions to each of the features. 
These “escape distances” can then be standardized to rank the features by local importance. 
RbX is guaranteed to satisfy a “sparsity” axiom,
which requires that features which do not enter into the prediction model are assigned zero importance. 
At the same time, real data examples and synthetic experiments show how RbX can more readily detect all locally relevant features than existing methods.
\end{abstract}

\section{Introduction}
Suppose we have a prediction model $\hat f(x)$ to estimate a scalar outcome $y$ given a set of features $x\in\R^d$. 
We do not assume that $\hat{f}$ takes any particular functional form.
Rather, we consider $\hat{f}$ as a black-box function to which we only have query acccess. 
That is, we may compute the value of $\hat{f}(x)$ for any desired input $x$,
but do not have any additional information about the structure of the model $\hat{f}$.

After making a prediction at a target point $x_0$,
we seek to quantify the local importance of each feature on the prediction.
Of those features involved, some will be more influential than others in the predictions from $\hat{f}$ near $x_0$.
We would like a systematic way of identifying these.

As a motivating example, consider a loan borrower applying for credit.
Statistical learning techniques are increasingly carried out to assess credit-worthiness of prospective applicants~\citep{chen2018FICO}.
For example, credit-granting agents may collect many features about the applicant including age, gender, occupation, etc., and use them to predict their probability of default.
If the bank denies their request, it is natural for the user to ask: \textit{Are there any critical features that the applicant could change to improve their outcome?}
We return to this setting in Section~\ref{section:FICO}, with a real data example based on the FICO Explainable Machine Learning Challenge.

We distinguish our problem,
which we call \emph{local prediction importance},
from the questions of feature selection and feature importance. 
Feature selection methods, such as the LASSO for linear models~\citep{tibshirani1996lasso} and modern extensions like LassoNet for black-box models~\citep{lemhadri2021lassonet}, 
aim to select a small subset of features to generate a predictive model with greater accuracy and/or interpretability.
In our setting, the prediction model $\hat{f}$ is given,
and we seek only to faithfully explain the predictions of that model,
without regard to the unknowable data-generating process that created the features and response.
Feature importance methods include popular permutation-based approaches introduced by~\citet{breiman2001rf} for random forests,
which were extended to generic black-box models by~\citet{fisher2019all} and to a local method by~\citet{casalicchio2018}.
These approaches fix the prediction model $\hat{f}$,
but provide importance measures based on changes in the predictive performance of that model,
and thus also depend on the data-generating process.
By contrast, the term \emph{prediction importance} emphasizes the singular role of the structure of prediction model $\hat{f}$,
independent of how the model was trained or validated.

The distinction between local prediction importance and local feature importance is not always made in the literature.
However, it is relevant for a user who only cares about understanding the output of a given black-box model,
and does not want prediction explanations conflated with the underlying signal the model is trying to approximate.

\section{Why region-based explanations?}
Our proposed approach to local prediction importance is via region-based explanations (RbX). 
RbX is ``model-agnostic," meaning it does not require any knowledge about the structure of $\hat{f}$.
We defer a detailed description of the algorithm to Section~\ref{sec:alg},
but given a target point $x_0$,
the main idea is to construct a polytope that approximates the region in feature space with prediction values ``close`` to 
the prediction at $x_0$.
The user can define this ``closeness region" in a way that is relevant for the context in which the model is being deployed.
For example, if $\hat{f}$ corresponds to estimated class probabilities in a binary classification setting,
the region could consist of all prediction values on the same side of the decision boundary.
If $\hat{f}$ predicts a numeric medical outcome, and $x_0$ corresponds to a healthy patient,
then the region might be the accepted range of healthy outcomes.

Given this polytope approximation of the closeness region,
we then argue that distances from $x_0$ to the boundaries of this polytope in directions parallel to the coordinate axes inform the local sensitivities of $\hat{f}$ to each feature in a desirable way.
The approach is perhaps best motivated when the features are continuous,
but works for ordered categorical features (including binary features) as well.

\subsection{Previous work}
Existing approaches to local prediction importance can be broadly divided into two categories: surrogate methods and gradient-based methods.
Surrogate methods locally approximate $\hat{f}$ by fitting a simpler prediction model that treats the predictions of $\hat{f}$ in a region near $x_0$ as the response.
The weights assigned to each feature in this model are then used for local importance.
For instance, LIME \citep{ribeiro2016should} draws feature instances from a density centered at the target point $x_0$ and uses a linear surrogate.
\cite{lundberg2017unified} propose Kernel SHAP (hereafter just SHAP), which they showed is an algorithmic approximation to fitting an additive surrogate model
with weights corresponding to Shapley values.

Gradient-based methods consider infinitesimal regions on the decision surface and use the resulting first-order approximation to derive local feature importance. 
For example,~\citet{baehrens2010explain} provide local prediction importances based on the absolute value of the components of the gradient vector $\nabla \hat{f}(x_0)$;
their approach for estimating this gradient is by fitting a global surrogate model using Parzen windows.
Integrated gradients~\citep{sundararajan2017IG} considers the line integral of the components of the gradient of $\hat{f}$ over a straight line path in feature space from a baseline point $x$ to $x_0$.
Other gradient methods are not model-agnostic.
For instance, DeepLIFT \citep{shrikumar2017learning} relies on backpropagation to estimate gradients in neural networks.

\subsection{Sparsity and detection power}
\label{sec:dummy_behavior}
RbX is designed to satisfy two properties,
loosely analogous to type I error control and power in classical hypothesis testing.
\begin{ppty}
\label{ppty:sparsity}
(Sparsity) A feature not involved in the prediction model $\hat{f}$ is assigned no importance.
\end{ppty}
Sparsity says we don't want a local prediction importance method to make any ``false discoveries"
by asserting that a completely irrelevant feature is important.
Of course, sparsity is not \emph{sufficient} for a \emph{good} local prediction importance method.
We also want ``detection power":
\begin{ppty}
\label{ppty:detection_power}
(Detection power) Any locally relevant feature for $\hat{f}$ has nonzero importance.
\end{ppty}
Non-axiomatic local prediction importance methods such as LIME and L2X~\citep{chen2018learning} --- a method that computes local feature scores by maximizing a variational relaxation of the mutual information between $y$ and the features $x$ encoded by $\hat{f}$ ---
evaluate their methods based on sparsity and detection power.
For instance, the experiments in~\citet{ribeiro2016should} show that LIME does a better job than some baseline methods in finding the features used in sparse logistic regression models and decision trees.
~\citep{chen2018learning} show that L2X does better than LIME, SHAP, and various gradient methods in recovering the relevant features in a sparse signal approximated by a dense neural network.

For the purposes of local prediction importance (rather than local feature importance),
what matters is whether a method can recover the relevant features in a sparse \emph{prediction model} (rather than a sparse \emph{signal}).
Thus, our synthetic experiments in Section~\ref{section:experiments} replace the dense neural net from~\citet{chen2018learning}
with sparse models.

SHAP and IG satisfy sparsity axiomatically.
LIME and L2X do not.
Unlike sparsity, detection power is not precisely defined,
due to subjectivity in the definition of ``locally relevant".
But a reasonable sufficient condition for local relevance of feature $j$ is for the $j$-th component of $\nabla \hat{f}(x_0)$,
the gradient of the prediction at $x_0$,
to be nonzero.
Then in the case that $\hat{f}$ is a sparse, additive regression model,
Property~\ref{ppty:sparsity} corresponds to assigning zero importance to all features with zero coefficients,
while Property~\ref{ppty:detection_power} means assigning nonzero importance to all other features.
A simple gradient-based method using finite differences would then always perfectly satisfy both properties,
as the set of features with nonzero gradients would always be precisely the relevant features.
By contrast, LIME only does this 90\%-92\% of the time in the experiments from~\citet{ribeiro2016should}.

What remains is to improve detection power in nonlinear models
without sacrificing sparsity.
In such models, a feature might be locally relevant near the target point $x_0$,
even if $\nabla \hat{f}(x_0)=0$. 
~\citet{sundararajan2017IG} motivate IG in this way,
noting that gradient methods fail when $\hat{f}$ has zero gradient with respect to a particular feature,
but still varies in that direction within a non-infinitesimal neighborhood that is considered locally relevant.

IG addresses this zero-gradient issue by accumulating gradients along the entire line segment between some baseline feature combination $x$ and the target point $x_0$.
However, this still only detects features that happen to vary infinitesimally somewhere along this line segment.
There are a lot of additional areas of the feature space near $x_0$ where $\hat{f}$ could depend on a given feature.
Our approach, RbX, examines the sensitivity of $\hat{f}$ in a large number of directions,
while inheriting sparsity from the finite differences gradient method.
It does so in a non-infinitesimal neighborhood of $x_0$,
adapting the search space to cover the entire region the user deems locally relevant via the closeness region.

\subsection{Baseline features}

Many methods including IG and SHAP rely on the specification of a baseline feature combination $x$,
such that the sum of all feature explanation scores at any target $x_0$ is equal to $f(x_0)-f(x)$.
While~\citet{sundararajan2017IG} note that natural baselines exist in settings like image classification and sentiment analysis,
for a general prediction or classification setting there may not be a canonical choice, 
and the feature attributions will be sensitive to the choice of baseline.
SHAP's reliance on a baseline feature combination is eliminated by the cohort Shapley method of~\citet{mase2019explaining}, 
though cohort Shapley still retains an additivity constraint that all feature attributions must add up to $\hat{f}(x_0)-\bar{f}$,
where $\bar{f}$ is the mean prediction on a set of $n$ observations.
If $\bar{f}$ is not a meaningful value then the individual feature scores do not have a direct interpretation.

Instead of requiring the user to pick a single ``representative" combination of \emph{features},
RbX asks for a region of prediction values on the \emph{outcome} scale that are ``close" to the prediction for the target point $x_0$.
We believe this to be a simpler and more interpretable decision point in many settings,
particularly when there are complex interactions between the features.
Region-based explanations can also avoid using information from areas of feature space that are implausible (see Appendix~\ref{sec:trust_regions}).

\section{The RbX algorithm}
We now describe the details of RbX.
For ease of exposition we assume the closeness region is an interval $\mathcal{I} = [\hat{f}(x_0)-\epsilon_L, \hat{f}(x_0)+\epsilon_H]$,
which depends on the user's choice of nonnegative parameters $\epsilon=(\epsilon_L,\epsilon_H)$.
For example, if the prediction at the target point $x_0$ is $\hat{f}(x_0)=13$,
and the user thinks predictions between 10 and 20 should be close,
then $\epsilon_L=3$ and $\epsilon_H=7$.
We define $\mathcal{E} = \{x \in \mathbb{R}^d \mid x \in \mathcal{I}\}$
to be the feature values for which the predictions are ``$\epsilon$-close" to $\hat{f}(x_0)$.
Points outside $\mathcal{E}$ are said to be ``$\epsilon$-far."
The RbX algorithm approximates $\mathcal{E}$ by a polytope $\ccp$.

\begin{definition}
 \label{def:polytope}
 A{ polytope} $\mathcal{P} \subset \mathbb{R}^d$ is any finite intersection of affine halfspaces, i.e.
 $$\mathcal{P} \equiv \cap_{1 \leq k \leq K} H_k,$$ 
 where $H_k = \{x \in \mathbb{R}^d: x^Tu_k \leq c_k \}$ is defined by its normal vector $u_k \in \mathbb{R}^d$ and intercept $c_k \in \mathbb{R}$.
\end{definition}

The use of a polytope approximation,
as opposed to a smooth shape like an ellipsoid,
enables sparsity.

\subsection{A polytope approximation algorithm}
\label{sec:alg}
RbX is a greedy procedure that constructs the polytope approximation $\ccp$ of the $\epsilon$-close region $\mathcal{E}$ one halfspace at a time (Algorithm~\ref{alg:RbX}).
The algorithm requires a set of \textit{context samples} $\mathcal{X} = \{x_i\}_{1 \leq i \leq n}$
that form the basis of the sampling procedure.
These should be representative feature combinations,
for instance from a possibly unlabeled training or validation set.
To make the polytope $\ccp$ scale equivariant, all the features are scaled to have standard deviation 1 across the context samples.
Only $\epsilon$-far context points are used by the remainder of the procedure.

\begin{algorithm}
    \caption{The RbX Algorithm}
    \label{alg:RbX}
	\begin{algorithmic}[1]
	\State \textbf{Input:} 
	target $x_0 \in \mathbb{R}^d$,
	closeness thresholds $\epsilon \succeq 0$,
	prediction model $\hat{f}$ with query access,
	context samples $\{x_i\}_{1 \leq i \leq n}$, maximum number of splits $K$.
	\State Compute $s$, the vector of standard deviations of the $d$ features across the context samples $\{x_i\}_{1\leq i \leq n}$.
	\State Standardize $x_i \leftarrow \diag(s)^{-1}x_i$ for $i = 0,1,\ldots,n$. 
	\For{$i \in [1:n]$ if $x_i$ is $\epsilon$-far}
	    \State Shrink $x_i$ onto the $\epsilon$-decision boundary using line search: $\tilde{x_i} \leftarrow \text{Line-Search}(x_i,\hat{f},\epsilon, x_0)$
	\EndFor
    \State Initialize $\mathcal{R} \leftarrow \{ \tilde{x_i} \}$
    \State Initialize the set of support vectors $\mathcal{S} \leftarrow \emptyset$
    \State Initialize $k \leftarrow 1$
	\While {$\mathcal{R} \neq \emptyset$ and $k \leq K$}
	    \State Find $\displaystyle \tilde{x}^{(k)} \leftarrow \argmin_{\tilde{x} \in R} \norm{\tilde{x} - x_0}_2$
	    \State Estimate gradient of $\hat{f}$ at $\tilde{x}^{(k)}$: $g_k \leftarrow \text{Estimate-Grad} (\tilde{x}^{(k)}; \hat{f}, \delta, r, m)$
	    \State Compute halfspace: $H_k \leftarrow \{x \in \mathbb{R}^d \mid x^T g_k \leq (\tilde{x}^{(k)})^T g_k\}$
	    \State Update region: $\mathcal{R} \leftarrow \mathcal{R} \cap \intr\  H_k$
	    \State Update support vectors: $\mathcal{S} \leftarrow \mathcal{S} \cup \{ \tilde{x}^{(k)} \}$
	    \State Increment $k$: $k \leftarrow k + 1$
	\EndWhile
	\State
	\Return $\{ H_k \}_{1 \leq k \leq n}$, the collection of halfspaces defining the polytope $\ccp$
	\State Notation: Line-Search is defined in Algorithm~\ref{alg:line-search}. Estimate-Grad is given in Algorithm~\ref{alg:estimate-grad}.
	\end{algorithmic} 
\end{algorithm}



The first step in the RbX algorithm
is to shrink each (standardized and $\epsilon$-far) context point along the line segment in $\mathbb{R}^d$ connecting it to $x_0$
to a point on the $\epsilon$-boundary.
This can be done quickly to exponential accuracy via a standard bisection-based line search 
(see Algorithm~\ref{alg:line-search} in the Appendix for an example),
noting that $x_0$ is always $\epsilon$-close.

Next, RbX finds $x^{(1)}$, the context point whose shrunken counterpart $\tilde{x}^{(1)}$ is closest to $x_0$ in Euclidean distance.
The first halfspace $H_1$ of the polytope $\ccp$ is chosen to pass through $\tilde{x}^{(1)}$ 
and have normal vector equal to an estimate of the gradient of $\hat{f}$,
computed using finite differences (Algorithm~\ref{alg:estimate-grad})
at $\tilde{x}^{(1)}$.
This is motivated by the fact that in the case that $\hat{f}$ is differentiable,
$H_1$ is a first-order approximation of a level set of $\hat{f}$.
Finally, all shrunken context points outside the interior of $H_1$ are discarded,
and the process is iterated with the remaining shrunken context points until either $K$ halfspaces have been learned,
or there are no more shrunken context points remaining.

Note that $\tilde{x}^{(k)}$, the shrunken context point chosen on the $k$-th iteration of the algorithm,
lies outside the interior of the $k$-th halfspace $H_k$ by construction.
Thus the number of remaining shrunken context points decreases by at least one after each iteration,
and so the algorithm terminates in at most $n$ iterations.
In practice, far fewer iterations are often needed.
To further reduce computation, one can impose early stopping by specifying a maximum number of halfspaces $K < n$ to be learned.

The ``greedy" nature of the algorithm stems from choosing the closest shrunken context points first.
This helps enforce a better approximation of the parts of the decision boundary closer to $x_0$,
i.e. the ``locally relevant" ones.
It also tends to decrease the number of iterations needed before termination,
as there will typically be more shrunken context points on the opposite side of each learned halfspace.




\subsection{Finite differences gradient estimation and sparsity}
The gradients of $\hat{f}$ at the shrunken context points are estimated using finite differences (Algorithm~\ref{alg:estimate-grad}).
We allow the user to average gradient estimates at $m$ ``jittered" points that are the original points corrupted by a small amount of Gaussian noise.
This jittering is designed to smooth the gradient estimate when $\hat{f}$ has discontinuities on the $\epsilon$-boundary.
If $\hat{f}$ has no dependence on the $i$-th feature,
it is clear that the $i$-th component of $\nabla \hat{f}(x)$ will be 0 according to Algorithm~\ref{alg:estimate-grad},
regardless of the parameters used.
Note this would not be the case if Algorithm~\ref{alg:estimate-grad} were replaced by a smooth gradient estimator,
such as the Parzen windows used by~\citet{baehrens2010explain}.
Having zero gradient estimates for irrelevant features ensures that our local importance scores satisfy sparsity,
as described in Section~\ref{sec:escape_distances}.
All results in this paper are presented with parameters $r=0.01$, $\delta=0.1$, and $m=10$.
\begin{algorithm}
    \caption{Estimate-Grad Algorithm - Finite Differences}
    \label{alg:estimate-grad}
    \begin{algorithmic}[1]
        \State \textbf{Input:} 
    	point $x \in \mathbb{R}^d$,
    	prediction model $\hat{f}$, step size $\delta$, jitter radius $r$, number of jitter samples $m$ 
	    \For {$j \in [1:m]$} 
    	    \State Generate $z \sim r \cdot \mathcal{N}(0, I_d)$
    	    \State $v \leftarrow x + z$
    	    \State $g_i^{(j)} \leftarrow \frac{\hat{f}(v+\delta e_i)-\hat{f}(v-\delta e_i)}{2\delta}$, $i=1,\ldots,d$ 
	    \EndFor \\
	    \Return $\nabla \hat{f}(x) = \frac{1}{m} \sum_{j=1}^m (g_1^{(j)},...,g_d^{(j)})$
    \end{algorithmic}
\end{algorithm}
\subsection{Toy example}
\begin{figure}[htp]
    \includegraphics[width=\linewidth]{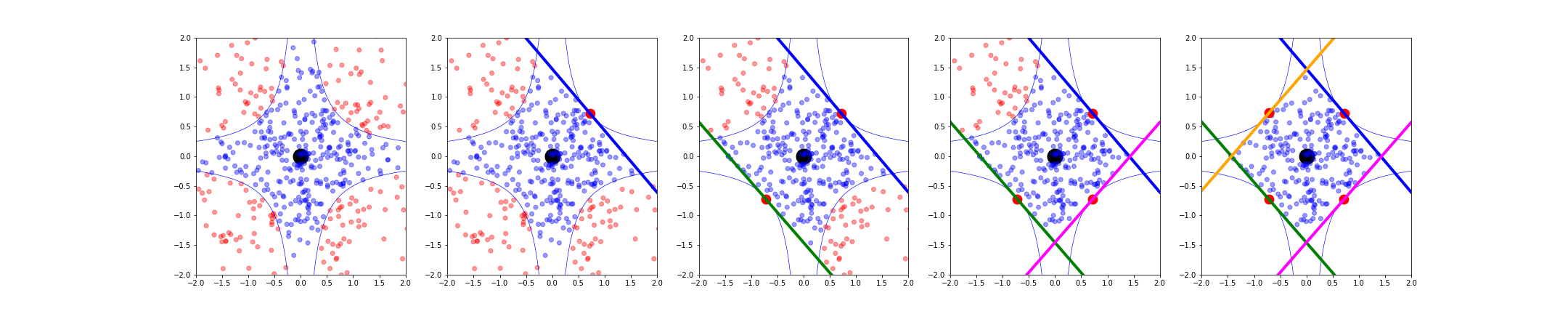}
    \caption{An illustration of the RbX algorithm for the pairwise interaction model $\hat{f}(x) = x_1 \cdot x_2$ with $\epsilon=(0.5,0.5)$ and target point $x_0=(0,0)$,
    highlighted in the center of each plot.
    The smaller dots are the context points, 
    colored by whether they are $\epsilon$-far.
    Each panel shows the additional halfspace constructed in one iteration of the RbX algorithm.
    The lines are the halfspaces learned at each step of the algorithm;
    the larger dots along these lines are the shrunken context points used.}
    \label{fig:XOR}
\end{figure}

We briefly illustrate the RbX algorithm in Fig.~\ref{fig:XOR},
in a toy example with $d=2$, $\hat{f}(x)=x_1 \cdot x_2$, $x_0=(0,0)$, and $\epsilon=(0.5,0.5)$.
500 context points generated from a standard bivariate Gaussian distribution were used.

Each iteration of the algorithm approximates a plane tangent to the $\epsilon$-boundary at the closest shrunken  context point.
All context points on the side of the plane not containing $x_0$ are then discarded.
The resulting $\ccp$ at termination (after 4 splits) is a diamond-shaped region that truncates the true $\epsilon$-close region,
which is non-convex and extends infinitely along the coordinate axes in both directions.

\section{From polytopes to local feature importance}
\label{sec:escape_distances}
Given the polytope $\ccp$ output by the RbX algorithm,
we derive local prediction importance scores $S_j(\ccp)$
using ``feature escape distances" for each feature $j \in \{1,\ldots,d\}$:
 \begin{align*}
     \centering
     S_j^+(\ccp) & = \inf \{\alpha > 0 \mid x_0+\alpha e_j \notin \mathcal{P}\}; \quad 
     S_j^-(\ccp) = \inf\{\alpha > 0 \mid x_0-\alpha e_j \notin \mathcal{P}\}; \quad \\
     S_j(\ccp) & = \min(S_j^+(\ccp), S_j^-(\ccp)) \cdot \sign(S_j^+(\ccp)- S_j^-(\ccp))
 \end{align*}
Here $e_j$ is the $j$-th standard basis vector in $\mathbb{R}^d$.
Then $S_j(\ccp)$ is the minimum signed distance needed to escape the RbX polytope $\mathcal{P}$
by varying only the $j$-th feature from the target point $x_0$.
If the corresponding ``escape path" goes through regions of feature space deemed unlikely or untrustworthy,
the $S_j$ can be set to $\infty$ (Appendix~\ref{sec:trust_regions}).

Note that the escape distances are reported on the original scales of each feature (before standardization),
which enables them to be interpreted individually without reference to the escape distances of the other features.
Alternatively, by reporting the escape distances on the standardized scale,
we can sort them (from smallest to largest in absolute value) to obtain a ranking of the local importance of the features (from most important to least important).

If $\hat{f}$ doesn't depend on a feature $x_j$,
then all of the halfspaces defining $\ccp$ will have normal vectors with 0 component in the $x_j$ direction.
This implies that the corresponding $S_j(\ccp)$ will be $\infty$,
and thus such features will have no importance,
showing our procedure satisfies sparsity.

It is natural to compare the feature escape distances $S_j(\ccp)$ with the ``simple feature escape distances" $S_j(\mathcal{E})$,
which use the original $\epsilon$-close region $\mathcal{E}$ 
in place of the polytope $\ccp$.
They can be computed via a line search similar to Algorithm~\ref{alg:line-search},
without running RbX.
Clearly, a feature ranking based on the $S_j(\mathcal{E})$ would also satisfy sparsity.
Furthermore, it would have better detection power than a gradient-based method.
This is because the $S_j(\mathcal{E})$ look beyond an infinitesimal neighborhood of $x_0$,
instead focusing on a typically larger region defined in terms of the prediction values of $\hat{f}$ to be meaningful.
However, it still cannot capture changes in $\hat{f}$ near $x_0$ that cannot be detected when only one feature is changed at a time from its value at $x_0$.

Since the polytope $\ccp$ looks in many directions around $x_0$,
using the RbX distances $S_j(\ccp)$ provides better detection power.
A simple example of this can be seen in Fig.~\ref{fig:XOR}.
There we have $\tilde{S}_1=\tilde{S}_2=\infty$ but $S_1\approx S_2 \approx 1.4$.
This is illustrated further by the data and experiments in the next section.

\section{Data example and synthetic experiments}
We now compare RbX to existing methods for local prediction importance on a credit scoring example alluded to in the introduction, along with simulated experiments.
\subsection{Credit scoring}
\label{section:FICO}
The home equity line of credit (HELOC) dataset from the FICO xML Challenge (\texttt{community.fico.com/s/xml})
contains the \texttt{RiskPerformance}
of 2,502 credit applicants.
This is a binary indicator of whether they were ever more than 90 days past due in the first two years after account opening.
The goal is to interpretably classify each individual as having either ``Bad" or ``Good" \texttt{RiskPerformance} based on 23 predictors that are all either quantitative or ordered categorical.
We consider the local prediction importance of a shallow and sparse decision tree.
The scalar prediction output is taken to be the predicted probability of a ``Good" \texttt{RiskPerformance}.

\begin{figure}
    \centering
    \includegraphics[width=15cm]{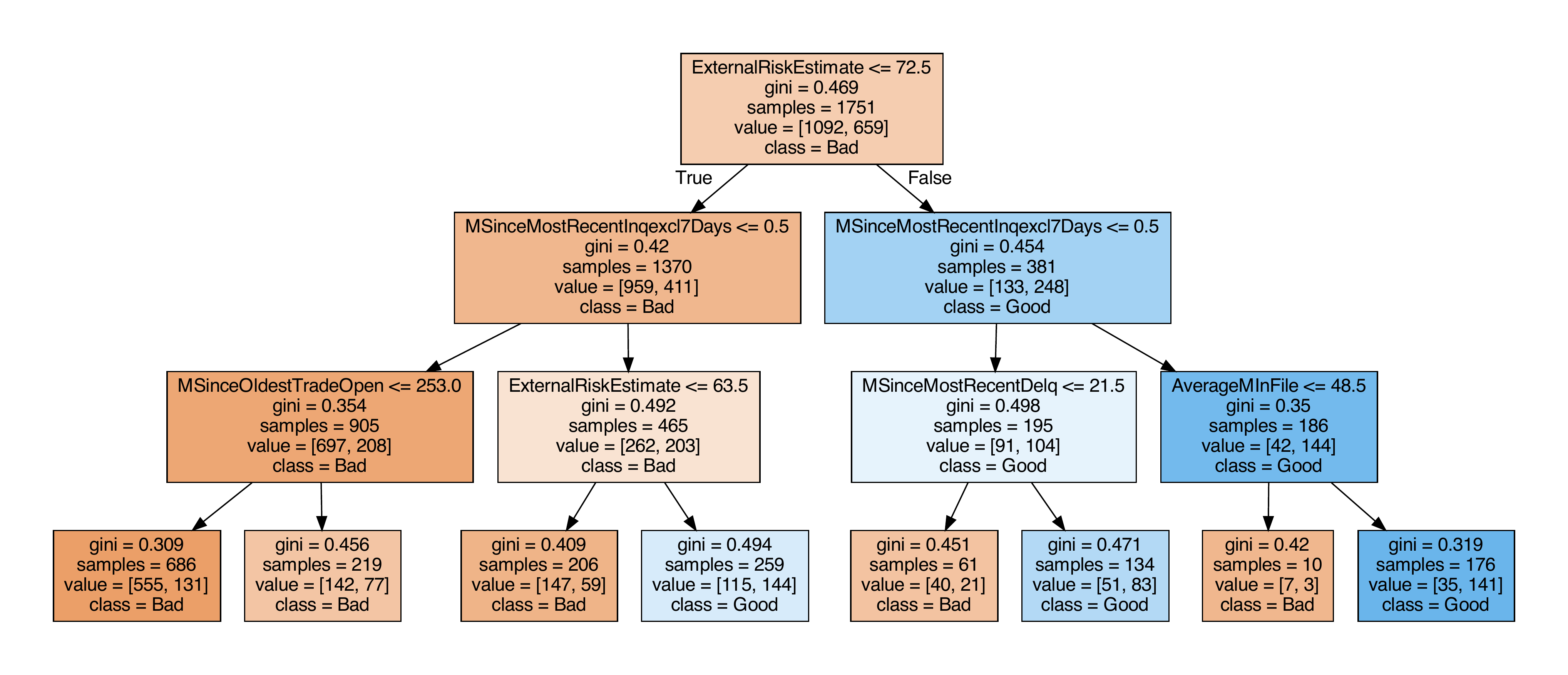}
    \caption{A visualization of a classification tree fit to the HELOC data from the FICO xML Challenge.}
    \label{fig:tree_viz}
\end{figure}

We split the dataset randomly into 1,751 training observations and 751 test observations.
The tree classifier is fit to the training observations and has depth 3.
A visualization of the classifier is in Fig.~\ref{fig:tree_viz}.
Without hyperparameter tuning,
it achieves an out-of-sample misclassification rate of around 30\%, compared to 26\% for the state of the art~\citep{chen2018FICO}.

We focus on a randomly chosen target point,
corresponding to the individual labeled 5,238 in the dataset.
Selected feature values for that individual (who has prediction value 0.191) are given in Table~\ref{table:FICO},
along with feature scores from various local importance methods.
For LIME and SHAP, we compute the scores from the open-source implementations and default settings,
except that for LIME we do not discretize the features,
which greatly improves its performance.
For RbX we assume a decision boundary of $\{x \mid \hat{f}(x)=0.5\}$,
and take $\mathcal{E}$ to be the set of points on the same side of this decision boundary as $x_0$.
All training observations are used as context points.

\begin{table}
\caption{\label{table:FICO} Feature values and importance scores for a sample target point based on $\hat{f}$ as in Fig.~\ref{fig:tree_viz}}
\centering
\begin{center}
\begin{tabular}{ccccccc}
\toprule
Feature &  Feature value & $S_j$ & $S_j$ norm. & LIME & SHAP 1 & SHAP 2 \\
\hline
\texttt{ExternalRiskEstimate} 
& 61.0 & 2.5 & 0.32 & 0.12 & -0.13 & -0.13  \\
\texttt{MSinceOldestTradeOpen}  & 149.0 & $\infty$  & $\infty$ & 0.018 & 0 & -0.08 \\
\texttt{MSinceMostRecentInqexcl7Days} 
& 0.0 & 0.58 & 0.12 &  0.084 & -0.23 & -0.15 \\
\texttt{MSinceMostRecentDelq} & 3.0 & $\infty$ & $\infty$ &  0.0067 & 0 & 0 \\
\texttt{AverageMInFile} & 49.0 & $\infty$ & $\infty$ &  0.016 & 0 & 0 \\
\texttt{NumTrades60Ever2DerogPubRec} & 1 & $\infty$ & $\infty$ & 0.0011 & 0 & 0
\end{tabular}
\end{center}
\end{table}

The RbX algorithm
terminates in 6 iterations.
We obtain escape distances
(on the original scale of the features) of 2.5 for \texttt{ExternalRiskEstimate} and 0.58 for \texttt{MSinceMostRecentInqexcl7Days}.
All other feature escape distances are infinite.
An intepretation is that increasing \texttt{ExternalRiskEstimate} by 2.5 and \texttt{MSinceMostRecentInqexcl7Days} by 0.58 would change the classification of the target individual under the tree classifier to \texttt{Good}.

By contrast, simple feature escape distances are infinite for \emph{every} feature,
as changing any single feature in the target point --- and keeping all others fixed --- cannot change the target individual's classification from Bad to Good under the tree model.
Once again,
this shows the increased detection power of the feature escape distances,
compared to the simple feature escape distances.

Table~\ref{table:FICO} also shows the sensitivity of SHAP explanations to the choice of baseline.
For two different baseline choices that are in the same leaf of the tree classifier,
SHAP gives different explanations.
One of these baselines yields nonzero SHAP value for \texttt{MSinceOldestTradeOpen},
yet using the other baseline gives zero SHAP for that feature
(Table~\ref{table:FICO}).
Finally, the LIME explanations for this example
seem to agree with RbX and SHAP,
except LIME assigns nonzero attribution scores to all six features that appear in the tree classifier.
Yet out of 1,000 LIME repetitions with random samples for fitting the surrogate linear model (one of which was randomly chosen to fill Table~\ref{table:FICO}),
46 of them yielded a higher attribution score to a feature that doesn't appear at all in the classifier
than to one of the six features that do,
demonstrating the failure of sparsity.

In Appendix~\ref{sec:boost} 
we also examine local prediction attribution scores for a gradient boosted tree ensemble.
Since it is difficult to visualize this classifier,
and it is not sparse,
there is not a clear ground truth we can use to evaluate the different methods' explanations.
However, we do note that as for the decision tree, all the simple feature escape distances are infinite,
so that method provides no information.
We provide a more systematic comparison of the different approaches in the synthetic experiments of the next section.

\subsection{Synthetic experiments}
\label{section:experiments}
To better illustrate the detection power of different local prediction importance methods,
we carry out a modification of the experiments of~\citet{chen2018learning},
with four data generating scenarios having sparse signals.
As explained in Section~\ref{sec:dummy_behavior},
we replace the dense neural network used there by two fully sparse prediction models.
In particular we evaluate performance on both the Bayes prediction model (smooth) and a K-nearest neighbors (KNN) regressor (nonsmooth).

The precise data scenarios are described as follows:
\begin{enumerate}
    \item Generate $(X_1,\ldots,X_9)$ from a spherical standard Gaussian distribution. Generate $X_{10}$ independently from an equally weighted mixture of two Gaussian distributions with standard deviation 1, centered at $+3$ and $-3$. 
    \item Let $X=(X_1,\ldots,X_{10})$. Then an outcome $Y$ is generated as follows:
    \begin{itemize}
        \item \underline{XOR:} $\E(Y \mid X=x) = (1+x_1x_2)^{-1} := p_X(x_1,x_2)$
        \item \underline{Orange skin:} $\E(Y \mid X=x) = \left(1+\exp\left(\sum_{i=1}^4 x_i^2 - 4\right)\right)^{-1} := p_O(x_1,x_2,x_3,x_4)$
        \item \underline{Nonlinear additive:} $\E(Y \mid X=x) = (1+\exp(-100\sin(2x_1) + 2|x_2| + x_3 + \exp(-x_4)))^{-1} := p_N(x_1,x_2,x_3,x_4)$
        \item \underline{Feature switching:} $\E(Y \mid X=x) = p_O(x_1,\ldots,x_4) r(x_{10}) + p_N(x_5,\ldots,x_8) (1-r(x_{10}))$
    \end{itemize}
\end{enumerate}
The feature switching scenario represents the setting where $Y$ is drawn from the orange skin model using features $X_1,\ldots,X_4$ whenever $X_{10}$ is from the component with center $+3$.
Otherwise, $Y$ is drawn from the nonlinear additive model with features $X_5,\ldots,X_8$.
Then $r(x_{10})$
is the posterior probability that $X_{10}$ was drawn from the component with center $+3$, given $X_{10}=x_{10}$.

As in~\citet{chen2018learning},
for the feature switching case we deem
features 5-9 are locally relevant if $x_{10}$ is nonnegative;
otherwise features 1-4 and 9 are locally relevant.
In the other scenarios,
the globally relevant features are the locally relevant ones for all target points.

For each method we select the $M$ locally most important features,
where $M$ is the true number of locally relevant features (2 for XOR, 5 for switch, 4 for the other two).
Ties in importance scores are broken randomly but features with no importance are never selected.
We evaluate the ability of RbX feature importance scores to recover  the locally relevant features under each scenario,
compared with popular methods with publicly available (or trivial) implementations:
simple feature importance (SFI), a gradient method using Algorithm~\ref{alg:estimate-grad} (``Gradient"), LIME, and SHAP. 
Of these approaches, only LIME does not satisfy sparsity.

We evaluate performances for each prediction model, scenario, and method on the same 1,000 randomly generated target points.
The KNN models are fit with $K=5$ using 1,000 independent training points
and only look at the locally relevant features (to ensure they are sparse).
For RbX we use 1,000 independent context points.
The $\epsilon$-close region for each target point consists of all points on the same side of the decision boundary $\{x \mid \hat{f}(x)=0.5\}$.
For SHAP the baseline point is the origin.

\begin{figure}
  \includegraphics[width=0.5\linewidth]{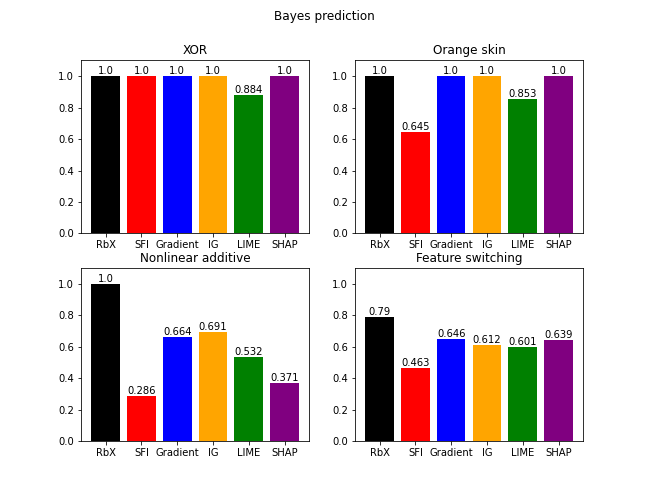}
  \includegraphics[width=0.5\linewidth]{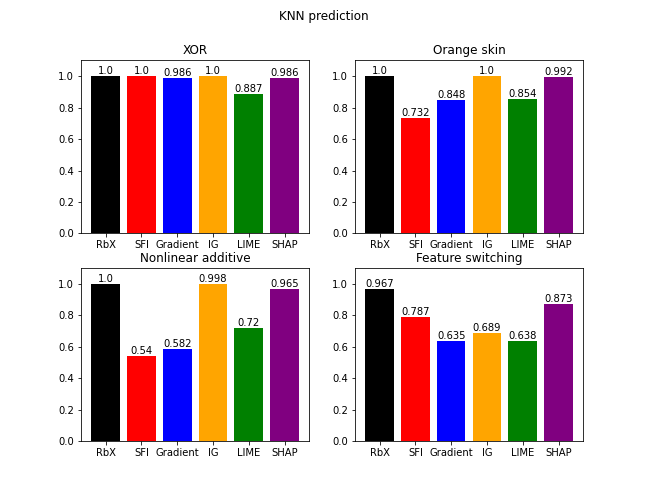}
    \caption{
     (Left) Recovery rates, based on 1,000 random target points, of the locally relevant features for various local prediction importance methods, using the smooth, sparse Bayes prediction model for each of the 4 simulated scenarios described in the main text.
     (Right) Same as the left panel, but for the nonsmooth KNN prediction model.
    }
    \label{fig:experiments}
\end{figure}

The results are shown in Fig.~\ref{fig:experiments}.
For all scenarios, RbX shows the highest recall among the methods considered,
with perfect performance on both classifiers for all scenarios besides feature switching.
The gradient-based and simple feature importance methods particularly struggle with the nonlinear additive model since there are regions where the classifiers are locally flat --- for instance, wherever $\sin(2X_1)$ is sufficiently greater than zero so that the predictions in a neighborhood are all numerically equivalent to 1.
RbX overcomes this by examining a non-infinitesimal neighborhood around $x_0$.

\section{Summary}
We have proposed region-based explanations (RbX) as a novel approach for generating instancewise explanations of black-box prediction models.
The method is agnostic to the inner workings of the prediction model,
only requiring query access to it. 
The main idea behind RbX is quite simple --- 
it successively refines polytope approximations to a region of feature space with similar predictions to the target point.
The user can directly specify what prediction values are ``similar" based on the context in which the prediction model is being used.
By contrast, existing methods specify a ``locally relevant" region in terms of the values of the \emph{features}.
When the number of features is moderate and there are interactions between them,
specifying such a region becomes difficult.

Unlike some other widely used methods,
RbX is guaranteed to preserve sparsity, 
meaning that features which are completely irrelevant for a prediction model are always assigned zero importance.
At the same time, our data examples and simulation results suggest RbX has a strong ability to detect locally relevant features,
particularly in sparse prediction models.

Further work in the direction of region-based explanations might leverage more theoretical work on polytopic approximation to develop mathematical guarantees about the output region~\citep{bronstein2008approximation,arya2012polytope}.
There may also be other ways in which the polytope constructed by Algorithm~\ref{alg:RbX} or a variant thereof could be useful for local prediction importance,
beyond the ``escape distances" of Section~\ref{sec:escape_distances}.
Finally,
we believe it could be informative to formalize the notion of detection power
and develop principled ways of maximizing it subject to sparsity.

\subsubsection*{Acknowledgments}
The authors thank Robert Tibshirani, Benjamin Seiler, and Hristo Paskov for helpful discussions that improved the quality of this manuscript.

\bibliography{main}
\bibliographystyle{abbrvnat}

\appendix
\section{Appendix}
\subsection{Line search algorithm}
Here we provide a sample line search algorithm for shrinking the context points given to the RbX algorithm to the $\epsilon$-boundary.
\begin{algorithm}
    \caption{Line-Search Algorithm}
    \label{alg:line-search}
    \begin{algorithmic}[1]
    \State \textbf{Input:} $\epsilon$-far point $x \in \mathbb{R}$ to be shrunk, thresholds $\epsilon \succeq 0$, prediction model $\hat{f}$, target $x_0 \in \mathbb{R}^d$, and maximum number of iterations $M$.
    \State Set $t_H \leftarrow 1$, $t_L \leftarrow 0$.
    \For {$\textnormal{iter} \in [1:M]$}
        \State Compute $t_M \leftarrow \frac{1}{2}(t_H+t_L)$
        \If {$x_0+t_M(x-x_0) \in \mathcal{E}$}
            \State $t_L \gets \frac{1}{2}(t_H+t_L)$
        \Else
            \State $t_H \gets \frac{1}{2}(t_H+t_L)$
        \EndIf
    \EndFor
    \State
    \Return $\frac{1}{2}(t_H+t_L)$ 
    \end{algorithmic}
\end{algorithm}
\subsection{Trustworthy regions}
\label{sec:trust_regions}
As discussed by~\citet{mase2019explaining},
a pitfall of baseline methods such as LIME, Kernel SHAP, and IG 
is that they often rely on predictions at implausible combinations of feature values, 
such as a graduation date before a birth date,
due to interactions between features.
Local prediction explanations utilizing such information have questionable fidelity.

To prevent the feature escape distances $S_j(\ccp)$ from using information about the classifier near implausible feature combinations,
we can establish a ``trustworthy region" $\mathcal{T}$ containing $x_0$ that corresponds to the set of plausible feature values.
Then if $S_j^{+}(\mathcal{T}) < S_j^{+}(\mathcal{P})$ --- meaning that in order to escape $\ccp$ by increasing the $j$-th feature from $x_0$, we must leave the trustworthy region $\mathcal{T}$ --- we set $S_j^{+}(\ccp)=\infty$.
We do the same thing for the $S_j^{-}(\ccp)$.
We could also make the simple feature escape distances $S_j(\mathcal{E})$ more trustworthy in the same way.


In some settings, domain knowledge informs a reasonable choice for $\mathcal{T}$.
Otherwise, there are many plausible ways to define $\mathcal{T}$,
assuming access to a large collection of plausible feature combinations,
such as the context points for the RbX algorithm.
One such method is given in Appendix 5 of~\citet{mase2019explaining}.
Another, based on Section 14.2.4 of~\citet{hastie2009ESL},
would be to estimate $r(x) = \frac{g(x)}{g_0(x)}$ where $g(\cdot)$ is viewed as an unknown joint density of the data generating process for the context points,
and $g_0(\cdot)$ is a known baseline density that is positive at each context point,
e.g. uniform over a rectangular region containing the context points.
This function $r(x)$ can be estimated by any binary classification procedure that outputs class probabilities.
By generating a large number of i.i.d. points from $g_0(\cdot)$, 
we can learn the probability that a point at $x$ came from $g$ rather than $g_0$,
using both the original context points and the feature combinations generated from $g_0(\cdot)$.
From this, an estimate $\hat{r}(x)$ of $r(x)$ follows via Bayes' rule.
Then we define $\mathcal{T} = \{x \mid \hat{r}(x) \geq \beta\}$ for some trustworthiness threshold $\beta >0$.

\begin{figure}[htp]
    \centering
    \includegraphics[width=\linewidth]{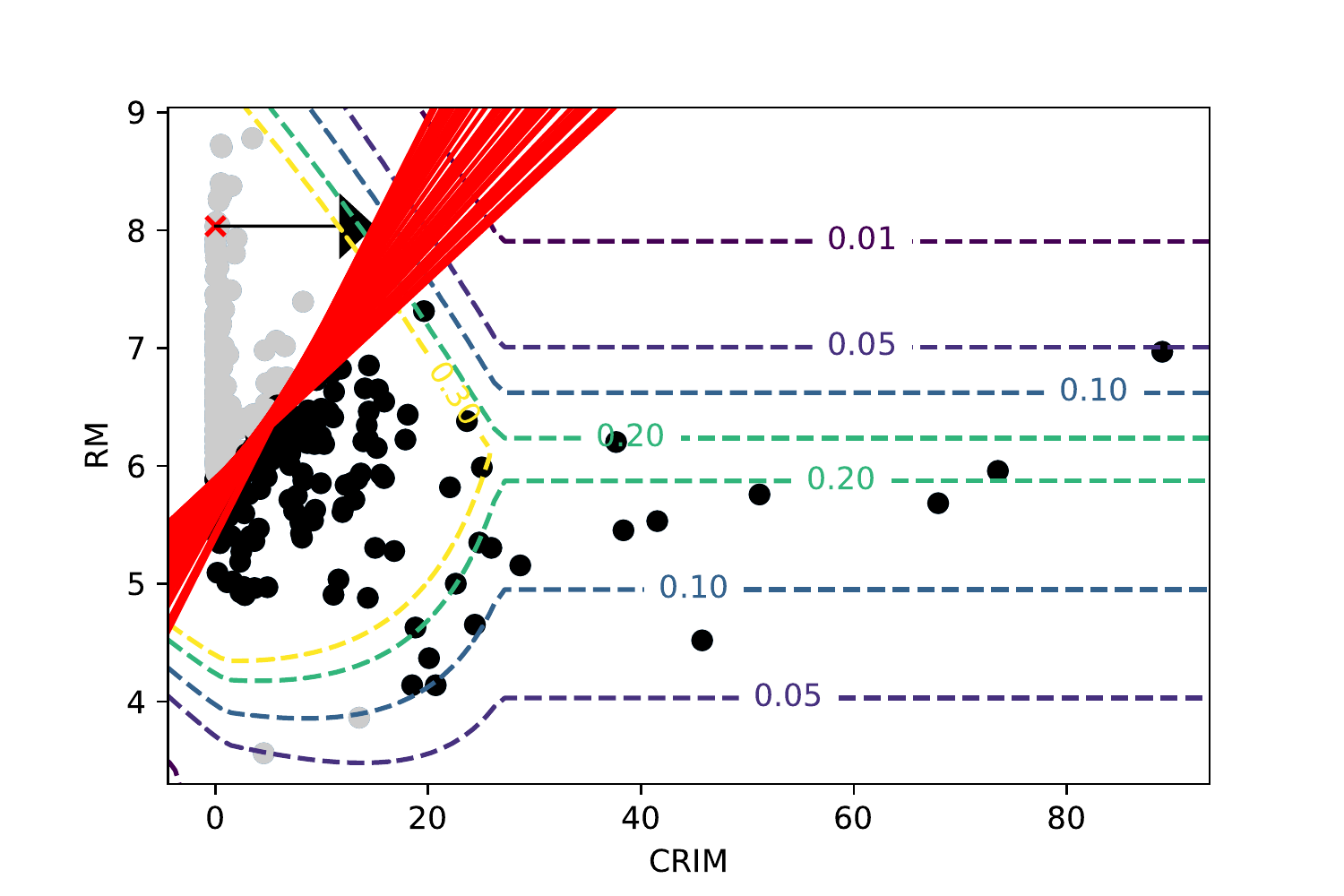}
    \caption{An illustration of the trustworthiness region classifier for the Boston housing data example described in the text. Its contours are indicated by the dashed lines. The solid lines indicate the halfspaces defining the polytope $\ccp$ from the RbX algorithm for the target point $x_0$, denoted with a red X. The arrow indicates the ``polytope escape path" from $x_0$ in the \texttt{CRIM} direction. For any $\beta$ such that the arrow crosses the $\beta$-contour (or smaller) of the trustworthiness region classifier, we set
    $S_{\texttt{CRIM}}^+(\ccp)$ to $\infty$. The dots are candidate context points from the entire dataset, colored by whether they are $\epsilon$-far.}
    \label{fig:boston}
\end{figure}

We illustrate a trustworthy region using a simple quadratic logistic regression classifier fit to the popular Boston housing dataset~\citep{harrison1978boston}.
The goal, as in~\citet{mase2019explaining}, is to predict whether median neighborhood value is less than \$20,000 based on two features --- \texttt{CRIM} (crime rate) and \texttt{RM} (median number of rooms per home).
We use the same target point as~\citet{mase2019explaining}, marked with a red X in Fig.~\ref{fig:boston}.

The trustworthiness classifier was fit with multivariate adaptive regression splines~\citep{friedman1991MARS} using a logistic link,
allowing for order-two interactions.
The contours of this classifier are shown in Fig.~\ref{fig:boston}.
If $\mathcal{T}$ is defined with any trustworthiness threshold $\beta$ larger than about 0.15,
the escape distance in the positive \texttt{CRIM} direction is set to $\infty$ because then it is impossible to escape the RbX polytope without leaving $\mathcal{T}$. 
As~\citet{mase2019explaining} suggest, this is desirable since there are no context points with similar $\texttt{RM}$ values in the $\epsilon$-far region but high values of \texttt{CRIM}.

\subsection{Results for boosted tree ensemble on FICO dataset}
\label{sec:boost}
We replicate the results given in Table~\ref{table:FICO},
but replacing the tree classifier in Fig.~\ref{fig:tree_viz} with a gradient boosted ensemble fit using XGBoost~\citep{chen2016xgboost}
on the same training dataset.
The training loss is the negative logistic log likelihood with early stopping after 10 iterations (and otherwise up to 1,000 boosting rounds).

\begin{table}[!h]
\label{table:FICO_boost}
\caption{
Same as Table~\ref{table:FICO}, but for the boosted tree ensemble classifier}
\centering
\begin{center}
\begin{tabular}{ccccccc}
\toprule
Feature &  Feature value & $S_j$ & $S_j$ norm. & LIME & SHAP 1 & SHAP 2 \\
\hline
\texttt{ExternalRiskEstimate} 
& 61.0 & -4.95 & -0.64 & -0.10 & -0.17 & -0.24  \\
\texttt{MSinceOldestTradeOpen}  & 149.0 & -15.7  & -0.17 & -0.045 & -0.12 & 0 \\
\texttt{MSinceMostRecentInqexcl7Days} 
& 0.0 & 1.72 & 0.40 &  -0.10 & -0.32 & -0.18 \\
\texttt{MSinceMostRecentDelq} & 3.0 & 8.2 & 0.39 &  -0.06 & -0.026 & 0 \\
\texttt{AverageMInFile} & 49.0 & -11.5 & -0.43 & -0.05 & 0 & 0 \\
\texttt{NumTrades60Ever2DerogPubRec} & 1 & 1.63 & 1.11 & 0.015 & 0.024 & 0.031
\end{tabular}
\end{center}

\end{table}

\end{document}